%% file: main.tex
\documentclass[journal,twoside,web]{ieeecolor}
\usepackage{tmi}
\usepackage{amsmath,amssymb,amsfonts}
\usepackage{algorithmic}
\usepackage{algorithm}
\usepackage{graphicx}
\usepackage{textcomp}
\usepackage{lipsum}
\usepackage{xcolor}
\usepackage[style=ieee]{biblatex}

\newcommand{\ie}{\emph{i.e.}~}
\newcommand{\cf}{\emph{cf.}~}
\newcommand{\etal}{\emph{et al.}}

\def\BibTeX{{\rm B\kern-.05em{\sc i\kern-.025em b}\kern-.08em
    T\kern-.1667em\lower.7ex\hbox{E}\kern-.125emX}}
\begin{document}
\title{ROAM: Random Layer Mixup for Semi-Supervised Learning in Medical Imaging}
\author{Tariq Bdair, Benedikt Wiestler, Nassir Navab, and Shadi Albarqouni
\thanks{T.B. is financially supported by the German Academic Exchange Service (DAAD). S.A. is supported by the PRIME programme of the German Academic Exchange Service (DAAD) with funds from the German Federal Ministry of Education and Research (BMBF). We thank Alex Lamb for his feedback on employing ROAM with skip connection}
\thanks{T. Bdair, N. Navab, and S. Albarqouni are with the Chair for Computer Aided Medical Procedures, Technical University of Munich, 85748 Munich, Germany (e-mail: t.bdair@tum.de).}
\thanks{Benedikt. Wiestle is with Department of Neuroradiology, Technical University of Munich, 81675 Munich, Germany.}
\thanks{N. Navab is also with the Whiting School of Engineering, Johns Hopkins University, Baltimore, MD 21218 USA.}
\thanks{S. Albarqouni is also with Biomedical Image Computing, Computer Vision Laboratory, ETH Zurich, Switzerland.}}

\maketitle

\input{abstract}


\input{introduction}
\input{method}
\input{experiments}
\input{discussion}
\input{conclusion}
\printbibliography
\end{document}

%% file: abstract.tex
\begin{abstract}
Medical image segmentation is one of the major challenges addressed by machine learning methods. Yet, deep learning methods profoundly depend on a large amount of annotated data, which is time-consuming and costly. Though, semi-supervised learning methods approach this problem by leveraging an abundant amount of unlabeled data along with a small amount of labeled data in the training process. Recently, MixUp regularizer has been successfully introduced to semi-supervised learning methods showing superior performance. MixUp augments the model with new data points through linear interpolation of the data at the input space. We argue that this option is limited. Instead, we propose ROAM, a \textit{random layer mixup}, which encourages the network to be less confident for interpolated data points at randomly selected space. ROAM generates more data points that have never seen before, and hence it avoids over-fitting and enhances the generalization ability. We conduct extensive experiments to validate our method on three publicly available datasets on whole-brain image segmentation. ROAM achieves state-of-the-art (SOTA) results in fully supervised (89.5\%) and semi-supervised (87.0\%) settings with a relative improvement of up to 2.40\% and 16.50\%, respectively for the whole-brain segmentation. 
\end{abstract}

%% file: introduction.tex
\section{Introduction}
\label{sec:introduction}
Medical image segmentation plays a fundamental role in the medical field since it provides a tool to examine different diseases and quantify the human organs \cite{sharma2010automated}. Nevertheless, the manual segmentation is a tedious task and requires highly experienced physicians and subject to intra/inter- observer variability \cite{hu2001automatic}. Therefore, finding a fully automated segmentation approach is of high importance to tackle such challenges. Recently, deep learning-based methods have achieved state-of-the-art performance in medical image segmentation~\cite{coupe2019assemblynetFull,gu2019net,roy2019quicknat}. One major drawback of this approach is that the necessity for a huge amount of annotated data which is oftentimes not available in medical imaging.
Fortunately, semi-supervised learning (SSL) framework provides the tool to alleviate this problem by utilizing a huge amount of unlabeled data along with a few annotated ones in intelligent and efficient ways. SSL methods can be categorized into four main categories; (i) consistency regularization, (ii) entropy minimization, (iii) generative model, and (iv) graph-based methods. Next, we briefly introduce these methods with a focus on SSL works with Medical Imaging.

\textit{Consistency Regularization.}
The focus of these methods is to train the model to predict the same output for different perturbations or augmentations of the input data. Mean-Teacher, one of the most successful method of consistency regularization, has been employed by \cite{cui2019semi} for brain lesion segmentation. They simply introduce a segmentation consistency loss to minimize the discrepancy between the outputs of unlabeled data under different perturbations. Similar approach was utilized by \cite{bortsova2019semi} for Chest X-ray images segmentation. Yet, \cite{yu2019uncertainty} included the uncertainty information to enable the student model to learn from the reliable targets for left atrium segmentation. \cite{li2020transformation} utilized transformation-consistent to enhance the regularization on the pixel-level. Interesting results were demonstrated on skin lesion, optic disk, and liver segmentation.

\textit{Entropy Minimization} 
forces the decision boundary to pass through low-density regions to minimize the entropy of the predictions. 
One way to achieve this, in SSL setting, is to generate pseudo labels for the unlabeled data using a model trained on the labeled data. Then the training process is repeated using both labeled and pseudo-labeled data~\cite{chapelle2009semi}. This approach has been employed by~\cite{bai2017semi} for cardiac image segmentation, where the pseudo labels were additionally fine-tuned using the conditional random field method. 
Close to the pseudo labeling is Co-Training where confident predictions from separate models, trained using different views of the data, are utilized to enhance the training. \cite{xia2020uncertainty} utilized Co-Training by enforcing multi-view consistency of the unlabeled data for the pancreas and multi-organ segmentation.

\textit{Generative Models} have been extensively used in the past few years to estimate the density distribution of the data using the concept of adversarial learning~\cite{goodfellow2014generative}. Specifically, two networks were used in the training process namely the generator and the discriminator networks. The goal of the generator is to produce fake data with high quality as the original data, while the goal of the discriminator is to distinguish between the fake and the original data. This idea has been utilized by \cite{zhang2017deep} for gland image segmentation by encouraging the discriminator to distinguish between the segmentation results of unlabeled and labeled images while encouraging the segmenter (generator) to produce results fooling the discriminator. \cite{nie2018asdnet} utilized attention-based approach, based on the confidence map from the confidence network (discriminator), to include the unlabeled data in the adversarial training for pelvic organs segmentation. \cite{chen2019multi} encouraged the model to learn discriminative features for segmentation from unlabeled images, using autoencoder trained to synthetic segmentation labels, to segment tumor and white matter hyperintensities in the brain.

\textit{Graph-Based} 
methods represent the data, both labeled and unlabeled, in a graph structure, where the nodes represent the data points, the edges represent the connectivity, and the weights represent the distance. Graphs can be used then to propagate the labels from the labeled data to the unlabeled ones based on the connectivity and similarity.  
\cite{baur2017semi} introduced this concept a regularization term to the main objective function for MS Lesion Segmentation. The term based on the Laplacian graph and attempts to minimize the distance between similar unlabeled and labeled data points in the hidden space. \cite{ganaye2018semi} took the advantages of the invariant nature of the brain structure to build an adjacency graph of the brain structures acting as a constraint to refine the predicted segmentation of the unlabeled data. 

Modern regularization methods such as Input MixUp \cite{zhang2017mixup}, and Manifold mixup \cite{verma2019manifold} have been recently introduced to avoid over-fitting by encouraging the model to be less confident for interpolated data points at the input space or the latent space respectively. Both methods have been successfully employed for fully supervised segmentation frameworks; e.g. cardiac image segmentation~\cite{chaitanya2019semi}, brain tumor segmentation~\cite{eaton2018improving}, knee segmentation~\cite{panfilov2019improving}, and prostate cancer segmentation~\cite{jung2019prostate}. These works have shown the effectiveness of MixUp over standard data augmentation methods in medical imaging. 
Recently, MixMatch \cite{berthelot2019mixmatch}, that inspired our work, introduced Input MixUp to the SSL paradigm achieving SOTA results in image classification. MixMatch augments the model with interpolated data at the input space. While this approach is interesting and indeed provides the model with diverse data points, it is rather limited. We argue that performing the mixup operation at \textit{randomly} selected hidden representations provides the network with novel representation and additional training signal that suits the complexity of medical image segmentation tasks. Our method takes the advantages of both MixMatch and Manifold mixup to boost the performance of the model leading to better generalizability.

Thus, our \textbf{contributions} can be listed as follows; 
\begin{itemize}
    \item First, we introduce the \textit{RandOm lAyer Mixup (ROAM)} to medical image segmentation problem in semi-supervised setting, that has never been seen before. ROAM overcomes the limitations of MixMatch by encouraging the network to be less confident for interpolated data points at randomly selected layer, hence reducing over-fitting and generalizing well to unseen data. 
    \item Second, we perform a comprehensive ablation study showing the importance of our design choices. Further, we discuss employing the manifold mixup with the presence of skip-connections in U-Net like architectures. 
    \item Third, we perform extensive experiments, following the recommendations of \cite{oliver2018realistic}, in evaluating our method under the presence of domain shift, class mismatch, and different amounts of un-/labeled data. 
    \item Fourth, we empirically show the effectiveness of ROAM by demonstrating a SOTA performance in both supervised and semi-supervised settings in the Whole Brain Image segmentation. 
    \item Fifth, we utilized a unified architecture for implementing different SSL methods for the sake of fair comparison. The code is made publicly available for benchmarking\footnote{\url{https://github.com/tbdair/ROAM}}.
    
\end{itemize}

%% file: method.tex
\section{Methodology}
\label{sec:methodology}
We first give an introduction to SSL Paradigm, Input MixUp, and Manifold MixUp, then introduce our method, namely ROAM.
\subsection{SSL Paradigm}
\label{sec:SSLParadigm}
In SSL, we are given a set of labeled $\mathcal{S}_L = \{ \mathcal{X}_{L},  \mathcal{Y}_{L} \}$ and unlabeled data $\mathcal{S}_{U} = \{ \mathcal{X}_{U}\}$, where $\{ \mathcal{X}_{L},  \mathcal{X}_{U} \} = \{\textit{\textbf{x}}_1,\ldots, \textit{\textbf{x}}_L, \textit{\textbf{x}}_{L+1},\ldots, \textit{\textbf{x}}_{L+U}\}$ are input images, $\textit{\textbf{x}}\in \mathbb{R}^{H\times W}$, and $\mathcal{Y}_{L} = \{\textit{\textbf{y}}_1,\ldots,\textit{\textbf{y}}_L\}$ are the segmentation maps, $\textit{\textbf{y}}\in \mathbb{R}^{H\times W \times C}$, for $C$ organs. Our goal is to build a model $\mathcal{F}(\textit{\textbf{x}};\Theta)$ that takes input image $\textit{\textbf{x}}$ and outputs its segmentation map $\hat{\textit{\textbf{y}}}$. 
To leverage both labeled and unlabeled data, the objective function takes the form
\begin{equation} 
\label{generalLoss}
\mathcal{L}_{Total} = \mathcal{L}_{Supervised}+ \beta \mathcal{L}_{Unsupervised},
\end{equation}
where $\mathcal{L}_{Supervised}$ denotes the supervised loss and trained using labeled data $\mathcal{S}_L$, $L_{Unsupervised}$ denotes the unsupervised loss and trained on both labeled $\mathcal{S}_L$, and unlabeled data $\mathcal{S}_{U}$, and $\beta$ is a weighing factor that controls the contribution of the unsupervised loss. 
The unsupervised loss can have different forms depending on the employed SSL aforementioned approaches. In this section, we will focus on the consistency-regularization approach, where its goal is to minimize the distance between the feature representations of the input data point $x$ and its perturbed version $\hat{x}$. Formally, $L_{Unsupervised}=d(f_{\theta}(x)$, $f_{\theta}(\hat{x}))$, where $d(\cdot,\cdot)$ is a distance metric.

\subsection{Preliminaries}
\label{sec:methodBack}
\begin{figure}[t]
\centering
\includegraphics[scale=0.8]{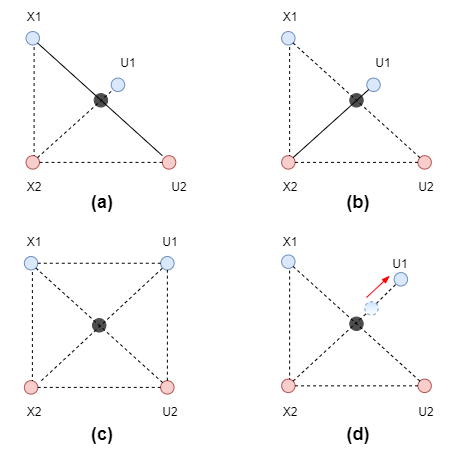}
\caption{Illustrative example. Fig. (a)-(b) Input Mixup: show the inconsistency of the generated soft label of grey-dot resulted from two different linear interpolations of the training examples. (c) Manifold Mixup: The learned hidden states are better organized and concentrated in local regions leading to the consistency of the soft labels. (d) Sharpening operation (red arrow) pushes the soft label of the unlabeled examples to a more confident region. Blue and Red colors represent different classes. Grey-dot represents the generated data point. Solid lines connect the interpolated data points.}
\label{fig0}
\end{figure}

\textit{Input Mixup} \cite{zhang2017mixup} is a simple data augmentation method that generates new data points $(x_{k}, y_{k})$ through a linear interpolation between a pair of training examples $(x_{i}, y_{i})$ and $(x_{j}, y_{j})$, 
\begin{align}
\label{eqMixup}
   x_{k} = \lambda x_{i}+ (1-\lambda) x_{j},\\ 
   y_{k} = \lambda y_{i}+ (1-\lambda) y_{j}, 
\end{align}
where $\lambda \in [0, 1]$. 
Mixup is considered as a type of data augmentation where the newly generated data points extend the training dataset following the cluster and manifold assumptions~\cite{chapelle2009semi} that linear interpolations of input examples should lead to linear interpolations of the corresponding labels. 
One major drawback of this approach is that the interpolations between two samples may intersect with a real sample leading to inconsistent soft-labels at interpolated points. Thus Input Mixup can suffer from underfitting and high loss. This can be better understood by examples. Fig.\ref{fig0}(a and b) show an illustrative example of generating the same data point (the grey-dot) through different ways of the linear combinations of labeled (X1 and X2) and unlabeled (U1 and U2) examples. In the first scenario, shown in Fig.\ref{fig0}.a,  the grey-dot is generated from combination of X1 and U2, hence, the generated soft label has an equal probability of blue and red classes (50\% each). In the second case,  the grey-dot is generated from combination of X2 and U1 with probability of 90\% blue class and 10\% red class (\cf Fig.\ref{fig0}.b). 

\textit{Manifold mixup} \cite{verma2019manifold}, on the other hand, overcomes the above limitations by training on the convex combinations of the hidden state representations of data samples. The learned hidden states lead to better organization of the representation for each class, where it more concentrated, and classes vary along distinct dimensions in the hidden space which capture higher level information and provide additional training signals. Thus, the inconsistency of soft-labels at interpolated points can be avoided, regardless of the interpolated data points (\cf Fig.\ref{fig0}.c). 
\subsection{ROAM}
\begin{figure}[t]
\centering
\resizebox{0.40\textwidth}{!}{
\includegraphics[scale=0.45]{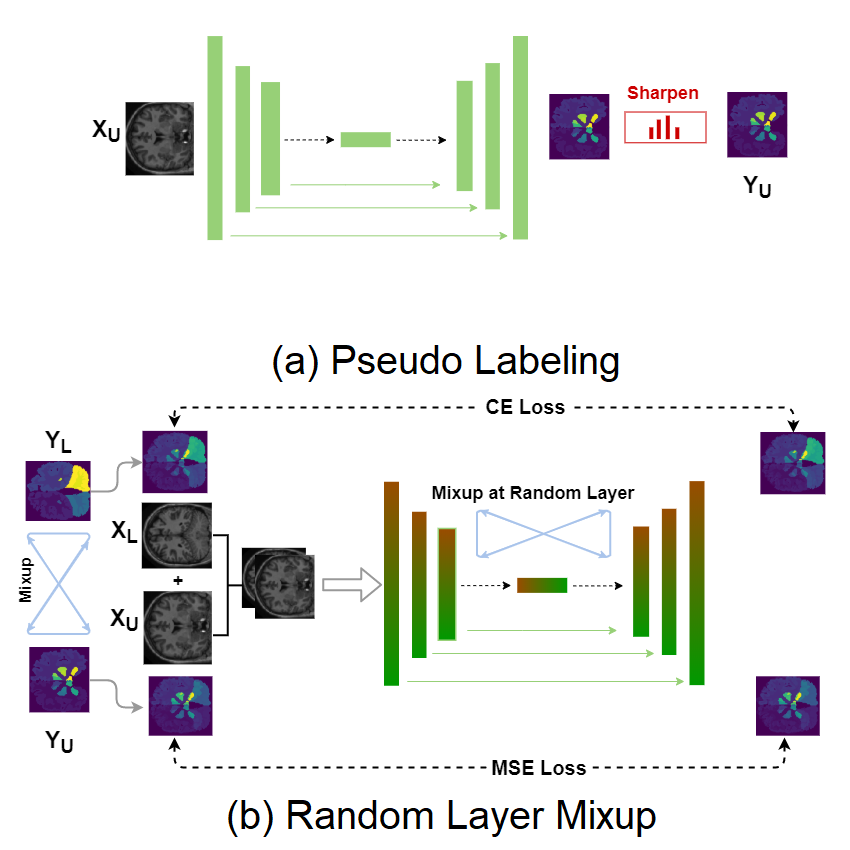}}
\caption{Illustration of our proposed method. (a) First, initial labels for the unlabeled batch is produced from a pre-trained model, then, a sharpening step is applied to fine-tune the labels. (b) Second, the labeled and unlabeled batches are concatenated before they fed to the network, and mixed at a random layer,\emph{e.g.,} Input layer in this figure.}
\label{fig1}
\end{figure}

The core components of our method are (a) \textit{Pseudo Labeling}: Given a pre-trained model for few epochs on labeled data, the initial labels for the unlabeled batch were produced, then refined by applying a sharpening operation. (b) \textit{Random Layer Mixup}: The labeled batch and the unlabeled batch were concatenated, then passed to the network as normal. Then, a mixup operation is applied at a random layer. At the same time, a mixup operation is applied to the corresponding labels. Finally, the process is continued from that layer to the output layer. In the following sections, we illustrate our methodology in detail, while the entire framework and the algorithm are shown in Fig.\ref{fig1} and Algorithm.\ref{alg:algo1}, respectively.
\subsubsection{Pseudo labels}
First, we leverage the unlabeled data along with the labeled one using two steps; i) sharpening the initial guess for unlabeled data to minimize its entropy following \cite{berthelot2019mixmatch}, and ii) mixup the labeled and unlabeled data at random layers following \cite{verma2019manifold}.
The unlabeled data are first fed to the model outputting the initial guess
\begin{equation} \label{eq1}
 \hat{\textit{\textbf{y}}}_{i} = \mathcal{F}(\textit{\textbf{x}}_{i}; \Theta);\hspace{10pt}\text{where}\hspace{10pt}  \textit{\textbf{x}}_{i} \in \mathcal{X}_{U},
\end{equation}
before post-processed by a sharpening operation, parameterized with $T$, which is highly inspired by the entropy minimization literature~\cite{berthelot2019mixmatch}. The pseudo label set is then defined as $ \mathcal{\tilde{Y}}_{U} = \{\tilde{\textit{\textbf{y}}}_{i},\ldots,\tilde{\textit{\textbf{y}}}_{U}\}$, where

\begin{equation} 
\label{eq2}
  \left. \tilde{\textit{\textbf{y}}}_{i} = \verb!Sharpening!(\hat{\textit{\textbf{y}}}_{i}, T) := \hat{\textit{\textbf{y}}}_{i} ^{\frac{1}{T}} \middle/ \sum _{j = 1}^{C} \hat{\textit{\textbf{y}}}_{j} ^{\frac{1}{T}} \right..
\end{equation}
Applying the sharpening operation to the guessed labeled produces more stable predictions through pushing the guessed labels away from the decision boundaries, specifically, to more confident regions. This effect can be easily seen in Fig.\ref{fig0}.d where the unlabeled data point U1 is moved closer to the right distribution. 
\subsubsection{Random Layer Mixup}
Given the unlabeled data $\mathcal{X}_U$ and its pseudo labels $\mathcal{\tilde{Y}}_{U}$, along with the labeled data $\mathcal{X}_L$ and its one-hot encoding labels $\mathcal{Y}_{L}$, we concatenate the two sets as $\mathcal{X}= \{\mathcal{X}_{L}, \mathcal{X}_{U}\}, \mathcal{Y}= \{\mathcal{Y}_{L}, \mathcal{\tilde{Y}}_{U}\}$. To enable running the mixup operation at randomly selected latent space, we first define $(\mathcal{H, Y})$, where
\begin{align} \label{eq3}
 \mathcal{H} =
\left\{
	\begin{array}{llllll}
	\mathcal{X},  & & & & & \kappa = 0 \\ 
	\mathcal{F}_{\kappa}(\mathcal{X}), & & & & & otherwise
	\end{array} 
\right.,
\end{align}
where $\mathcal{F}_{\kappa}(\cdot)$ is the hidden representation of the input data at layer $\kappa$. Note that, we select the input data when $\kappa = 0$.
To introduce a noisy interpolated data, a permuted version of the original data is created 
$\tilde{\mathcal{H}}, \tilde{\mathcal{Y}} = \verb!Permute!(\mathcal{H, Y})$,
and fed to the $\verb!MixUp!$ operation as
\begin{align} \label{eq4}
\mathcal{H}^\prime = \lambda^\prime \mathcal{H} + (1-\lambda^\prime) \tilde{\mathcal{H}},\\
\mathcal{Y}^\prime = \lambda^\prime \mathcal{Y} + (1-\lambda^\prime) \tilde{\mathcal{Y}},
\end{align}
where $\mathcal{H}^\prime$ and $\mathcal{Y}^\prime$ are the interpolated mixed-up data. 
To favour the original data over the permuted one, we set $\lambda^\prime = \mathrm{max}(\lambda, 1- \lambda)$, where $\lambda \in [0, 1]$ is sampled from a $\mathrm{Beta}(\alpha,\alpha)$ distribution with $\alpha$ as a hyper-parameter. Further, to keep the flow of the original data, we run some experiments where we do not run any mixup operation. In practice, we set $\kappa = 0$ and $\lambda^\prime = 1$, denoted as $\kappa = \Phi$. 
To this end, the mixed-up data $\mathcal{H}^\prime$ are fed to the model from layer $\kappa$ along the way to the output layer at which the segmentation maps are predicted $\mathcal{P}$. Eventually, $\mathcal{P}$ is split back into labeled and unlabeled predictions $\mathcal{P} = \{\mathcal{P}_{L}, \mathcal{P}_{U}\}$, and similarly $\mathcal{Y}^\prime$ into $\mathcal{Y}_{L}^\prime$ and $\mathcal{Y}_{U}^\prime$. 
\input{algorithem}

\subsubsection{Overall objective function}
is the sum of the cross entropy loss $\mathcal{L}_{CE}$ on the mixed-up labeled data, and the consistency mean squared loss $\mathcal{L}_{MSE}$ on the mixed-up unlabeled data, 

\begin{align} \label{eq9}
\arg\min_{\Theta} \mathcal{L}_{CE}(\mathcal{Y}_{L}^\prime, \mathcal{P}_{L})  +  \beta\mathcal{L}_{MSE}(\mathcal{Y}_{U}^\prime,\mathcal{P}_{U}), 
\end{align}
where $\beta$ is a hyper-parameter. 

%% file: algorithem.tex
\begin{algorithm}[t]
\caption{ROAM: Random Layer MixUp for SSL}\label{alg:algo1}
\begin{algorithmic}[1]
\REQUIRE{pre-trained model $\mathcal{F}(\cdot;\Theta^{(0)})$, labeled dataset $\mathcal{S}_{L}$, unlabeled dataset $\mathcal{S}_{U}$, batch size $B$, number of iteration $K$, The hyper-parameters $\{T, \alpha,\beta \}$}

\hspace{-15pt}\textbf{Initialize:} {$k \longleftarrow 0, \Theta \longleftarrow \Theta^{(0)}$}

\STATE \textbf{while} $k \leq K$ \textbf{do}

\STATE {\quad $\mathcal{B}_{L} \sim$  ($\mathcal{X}_{L}$, $\mathcal{Y}_{L}$);\hspace{10pt} $\mathcal{B}_{U}$  $\sim\mathcal{X}_{U}$}  \textcolor{gray}{\textit{//sample labeled and unlabeled batches}}

\STATE {\quad $\hat{\textit{\textbf{y}}}_{i} = \mathcal{F}(\textit{\textbf{x}}_{i}; \Theta)$; $x_{i} \in \mathcal{B}_{U}$} \textcolor{gray}{\textit{//initial labels for $\mathcal{X}_{U}$; Eq.\ref{eq1}}}

\STATE {\quad$\tilde{
\textit{\textbf{y}}}_{i} = \verb!Sharpening!(\hat{\textit{\textbf{y}}}_{i}, T)$} \textcolor{gray}{\textit{//pseudo labels; Eq.\ref{eq2}}}

\STATE {\quad$\mathcal{X}= \{\mathcal{X}_{L}, \mathcal{X}_{U}\}, \mathcal{Y}= \{\mathcal{Y}_{L}, \tilde{\mathcal{Y}}_{U}\}$}\textcolor{gray}{\textit{//concatenate both batches, $\mathcal{\tilde{Y}}_{U}$ from Eq.\ref{eq2}}}

\STATE {\quad$\kappa$ $\longleftarrow$ randomly select layer}

\STATE { \quad$\mathcal{H=F_{\kappa}(X)}$}\textcolor{gray}{\textit{//pass the data to the network, and extract $\mathcal{H}$; Eq.\ref{eq3}}}

\STATE { \quad$\tilde{\mathcal{H}}, \tilde{\mathcal{Y}} = \verb!Permute!(\mathcal{H, Y})$} \textcolor{gray}{\textit{//randomly shuffle the data}}

\STATE{\quad$\mathcal{H}^\prime, \mathcal{Y}^\prime=\verb!Mixup!(\alpha,\mathcal{H,Y,\tilde{H},\tilde{Y}})$} \textcolor{gray}{\textit{//perform mixup operation; Eqs.(\ref{eq4},8)}}

\STATE {\quad$\mathcal{P}$ $\longleftarrow$ resume passing $\mathcal{H}^\prime$ from layer $\kappa$ to the output layer}

\STATE {\quad$\mathcal{P_{L}, P_{U}} = \verb!Split!(\mathcal{P}); \mathcal{Y^\prime_{L}, Y^\prime_{U}} = \verb!Split!(\mathcal{Y^\prime})$} \textcolor{gray}{\textit{//split the predictions and labels}}

\STATE {\quad $\Theta \longleftarrow \arg\min_{\Theta} \mathcal{L}_{CE}(\mathcal{Y}_{L}^\prime, \mathcal{P}_{L})  +  \beta\mathcal{L}_{MSE}(\mathcal{Y}_{U}^\prime,\mathcal{P}_{U}$)} \textcolor{gray}{\textit{//calculate the loss; Eq.\ref{eq9}}}\\

\STATE {\textbf{end while}}
\end{algorithmic}

\end{algorithm}

%% file: experiments.tex
\section{Experiments}
\label{sec:experiments}
At first, we compare ROAM with SSL methods for medical images segmentation (Sec.~\ref{SecSSLResults}). The structures level results and qualitative results are presented (Sec.~\ref{SecStrctResults}) and (Sec.~\ref{SecQuaResults}) respectively. Followed by a comparison with SOTA methods for whole-brain segmentation in a fully-supervised fashion (Sec.~\ref{SecFSResults}). Then, we perform extensive experiments following the recommendations by \cite{oliver2018realistic} (Sec.~\ref{SecRE}), and investigate the performance of ROAM in the presence of domain shift (Sec.~\ref{secCrosD}). 
\subsection{Experiments Setup}
\subsubsection{Datasets}
We opt for three publicly available datasets for whole-brain segmentation; (i) MALC~\cite{landman2012miccai}, which consists of 30 T1 MRI volumes, 15 volumes splitted into 3 labeled ($\sim$500 slices), 9 unlabeled ($\sim$1500 slices), and 3 validation ($\sim$500 slices), and 15 testing volumes ($\sim$2500 slices), (ii) IBSR~\cite{rohlfing2011image}, which consists of 18 T1 MRI volumes ($\sim$2000 slices), and (iii) CANDI~\cite{kennedy2012candishare} which consists of 13 T1 MRI volumes ($\sim$1500 slices).
In all previous data settings, patient-wise random splitting strategy was considered in the training, validation, and testing data to avoid any overlaps. All images have a dimension of 256$\times$256 with a resolution ranges from $\sim$0.86 to 1.5mm, and the intensity values are normalized to $[0, 1]$.   
\subsubsection{Implementation details} We employ U-Net as backbone architecture. The weights are initialized with Xavier initialization, and trained using Adam optimizer. The learning rate, weight decay, and batch size are set to $0.0001$, $0.0001$, and $8$ respectively. The lower bound model is trained for 40 epochs, and the other semi-supervised models and the upper bound models are further trained for additional 40 epochs, where we pick the model with best validation accuracy. The hyper-parameters are set to $T=0.5$, $\alpha =0.75$, and $\beta =75$. The mixup layer $k$ selected randomly from the input, first, and last convolution layers denoted as $k=\{0, 1, L\}$. All the experiments performed on an NVIDIA GTX 1080 8GB machine. The training time is about 6 hours. PyTorch framework used for the implantation.
\subsubsection{Evaluation Metrics} 
We report the statistical summary of Dice score, in addition to the Hausdorff distance (HD), and the Mean Surface Distance (MSD). A Relative Improvement (RI) \emph{w.r.t} the baseline is also reported. 
\subsubsection{Baselines} 
One baseline is the initial model, denoted lower bound which trained on 3 labeled volumes. All SSL models  trained using the same 3 labeled data in addition to 9 unlabeled volumes. To compare SSL methods with fully-supervised model, we also define an upper bound, where the lower bound is further trained on the same additional 9 volumes, however, their labels are revealed. Note that, we use the MALC dataset for training and testing. For our method, we examine various choices for the mixed layers, among them MixMatch; \cite{berthelot2019mixmatch} when $\kappa = 0$, and manifold mixup when $\kappa = 2$.
\subsubsection{Regularized ROAM} In addition to the above, and in order to evaluate our contributions, we introduced our method (ROAM) as a regularizer to the fully supervised lower bound and upper bound models, denoted as ROAM-LB, and ROAM-UB, respectively.
\subsubsection{Models Selection}
\label{SecModSelc}
ROAM introduces the sharpening and concatenation to the manifold mixup. Also, it involves a set of hyper-parameters \ie $(\alpha, \beta)$ and design choices \ie $\kappa$ in its framework. Thus, for our model selection, we conduct an ablation study and sensitivity analysis for each of these parameters. We train our model for 80 epochs on the training dataset and select the model with the highest validation accuracy. The selected model will be used to report the results on the testing dataset. All results presented in Table \ref{table0}.   
\begin{table}[!t]
\centering
\caption{Mean Dice for Brain validation and testing datasets. ROAM, with $\kappa=\{0, 1, L\}$, sharpening, concatenation, $\alpha=0.75$, and $\beta=75$, obtains the best validation results, hence, will be our model selection. $\Phi$: no data mixup. All: all hidden layers. L: last layer.} \resizebox{0.5\textwidth}{!}{
\begin{tabular}{p{0.27\linewidth}p{0.27\linewidth}p{0.17\linewidth}p{0.13\linewidth}}
\hline
Ablation&Value&Validation&Testing\\
\hline
 \textbf{ROAM} &\{0, 1, L\}&\textbf{0.898}&0.870\\
\hline
$\kappa$    &0&0.881&0.852\\
            &1&0.867&0.843\\
            &2&0.894&0.872\\
            &3&0.868&0.825\\
            &4&0.863&0.828\\
            &5&0.877&0.847\\
            &L&0.865&0.843\\
            &\{0, 2, L\}&0.884&0.851\\
            &\{1, 2, L\}&0.883&0.863\\
            &\{0, 1, 5\}&0.881&0.860\\
            &\{$\Phi$, 0, 1, L\}&0.882&0.864\\
            &\{All\}&0.882&0.858\\
\hline
$\alpha$    &0.25&0.880&0.851\\
           &2&0.885&0.836\\
\hline
$\beta$    &0&0.893&0.844\\
\hline
Sharpening(\checkmark)    &Concatenation(\texttimes)&0.878&0.850\\
Sharpening(\texttimes)    &Concatenation(\checkmark)&0.861&0.823\\
Sharpening(\texttimes)    &Concatenation(\texttimes)&0.870&0.843\\
\end{tabular}}
\label{table0}
\end{table}
\paragraph{The selection of the random layer $\kappa$} 
\label{secAblKappa}
We conduct a sensitivity analysis on which layer(s) the mixup operation will obtain the best results. To do so, we examine a different set of layers including the input layer, any hidden layer, and the no-mix option where the original data passed to network as the usual training procedure. It can be seen from the results in Table \ref{table0} that there is a correlation between the selected layer and the validation results. Interestingly, this correlation is related to that mixing the data at different random layers achieves better results than using only one fixed layer except for $\kappa=2$. These results emphasize the importance of alternating the hidden space with the input space during the training process, which provides the model with novel variations of the data that never have seen using either the input or the hidden spaces.      
\paragraph{The concatenation \& the sharpening operations} 
\label{secAblSharp}
Turning to the concatenation \& the sharpening results. Specifically, we validate four combinations to study the effect of the sharpening and concatenation steps. The four combinations are (i) Sharpening step on the soft labels, then performing the mixup on a concatenated batch of labeled and unlabeled data. (ii) No sharpening step, then performing the mixup on a concatenated batch of labeled and unlabeled data. (iii) Sharpening step, then performing the mixup on a separated batch of labeled and unlabeled data. (iv) No sharpening step, then performing the mixup on a separated batch of labeled and unlabeled data. Overall, a drastic drop in the Dice score observed when removing one or both steps. Yet, the worse result obtained when mixing the data without applying the sharpening step. Together these results provide important insights into that mixing the soft labels without a sharpening step harms the quality of the labeled data.   
\paragraph{The hyperparameters $\alpha$ and $\beta$} 
\label{secAblalpha}
Moving to the hyperparameters, we examine three values of $\alpha=\{0.25, 0.75, 2\}$. $\alpha$ is set 0.75 as in \cite{berthelot2019mixmatch}, and set to 0.25 to favor one sample over the other, and set to 2 to make more balance between the different samples. Comparing the three results, it can be seen that selecting $\alpha=0.75$ makes the mixed-up data closer to the original data while maintaining the novelty of the generated points, hence, ROAM obtained better results. In the final part of our analysis, we examine two values of $\beta=\{0, 75\}$. $\beta$ is set 75 as in \cite{berthelot2019mixmatch} and set to 0 to evaluate the effectiveness of the newly generated data on the training without propagating the unlabeled loss. The results, as shown in Table \ref{table0}, show that ROAM make a use of the unlabeled loss effectively. Also, the obtained results at $\beta=0$ show that the random layer mixup operation boosts the performance of the model without the need unlabeled signal. These results represent a strong evidence that ROAM utilizes the unlabeled data efficiently in both cases. Together, the above analysis show that ROAM, with $\kappa=\{0, 1, L\}$, sharpening, concatenation, $\alpha=0.75$, and $\beta=75$, obtains the highest validation accuracy. Unless stated otherwise, we opt for these selections in the next experiments. Yet, in some experiments, we report the results for ROAM$(\kappa=0)$ to compare our method with MixMatch where the mixup performed at the input space. Also, we report the results of ROAM$(\kappa=2)$ as it obtains the second-highest validation accuracy and to evaluate our method at the manifold mixup. 
In summary, the results from previous analyses show the essential role of each component of our method on the segmentation tasks justifying its selection. Further discussion on the hyper-parameter tuning is presented in (Sec.\ref{Discussion}). Next, we move on to present the segmentation results on testing dataset.
\subsection{Segmentation Results}
\label{SecResultsBrain}
\subsubsection{Comparison with SSL methods} 
\label{SecSSLResults}
We compare our method with a set of SSL methods applied to medical imaging, namely Bai~\etal\cite{bai2017semi}, Baur~\etal\cite{baur2017semi}, Cui~\etal\cite{cui2019semi}, and Zhang~\etal\cite{zhang2017deep}. We opt for these methods based on the following criteria; first, we select one method from each of the SSL approaches, taking into consideration the ease of implementation, no additional losses introduced in the method, and the compatibility of the selected method with the unified architecture. Second, we rule out the 3D methods, and the ones introduce sophisticated training mechanisms such as Multi-view training, uncertainty estimations, and domain adaptation. Also, we exclude methods with no code available and difficult to implement and reproduce the results. 
Table \ref{table1} illustrates the results for whole-brain segmentation. It is apparent from this table that our method outperforms the lower bound and all previous works with a statistical significance ($p < 0.001$). 
The best results obtained by ROAM($\kappa=\{0, 1, L\}$) with reported average Dice of 87.0\% and RI about 16.50\%. Further analysis shows that ROAM($\kappa=\{0, 1, L\}$) outperforms its variant, \ie ROAM($\kappa=0$), which is similar to MixMatch. This can be justified by that our approach avoids over-fitting by introducing a lot of variations through generating novel data points that never seen before via random layer mixup operation. 
A similar performance is reported for ROAM($\kappa=2$).
Further statistical tests revealed that our method achieves the best HD and MSD with values of 3.87 and 1.00, respectively. 
Moreover, ROAM-LB and ROAM-UB models outperform their competitor significantly with average Dices of 82.3\% and 89.3\%, and RI of 10.17\% and 19.54\% respectively. This is a strong evidence that applying ROAM as a regularizer provides the model with new data points, hence, boosts the performance without the need of any additional data. 
Yet, the most interesting aspect of this experiment revealed when comparing ROAM-LB model with the other SSL methods. Except for Cui~\etal\cite{cui2019semi}, ROAM-LB outperforms all SSL methods by significant margins.   
\begin{table}[!t] 
\caption{Mean (Median) $\pm$ Std. of different evaluation metrics are reported on the MALC testing set for baselines and different SSL methods, including ours. *: significant improvement. L: Last layer. $\dagger$: MixMatch~\cite{berthelot2019mixmatch}. $\ddagger$:ROAM ($\kappa=\{0,1,L\}$). $\uparrow$($\downarrow$): The higher (lower) the better.} \label{table1}
\centering
\resizebox{0.5\textwidth}{!}{
\begin{tabular}{p{0.25\linewidth}p{0.25\linewidth}p{0.11\linewidth}p{0.14\linewidth}p{0.14\linewidth}}
\hline
Model Name & Dice Coefficient $\uparrow$ & RI(\%) $\uparrow$ & HD $\downarrow$ & MSD $\downarrow$\\
\hline
\hline
Lower Bound                       &0.747(0.769)$\pm$0.071*&0&4.16$\pm$0.43&1.06$\pm$0.088\\
\textbf{ROAM-LB}                       &\textbf{0.823(0.841)$\pm$0.052}&\textbf{10.17}&\textbf{4.07$\pm$0.35}&\textbf{1.05$\pm$0.071}\\
\hline
Bai~\etal\cite{bai2017semi}   &0.800(0.815)$\pm$0.055*&7.10&4.06$\pm$0.43&1.02$\pm$0.086\\
Zhang~\etal\cite{zhang2017deep} &0.819(0.851)$\pm$0.060*&9.64&4.02$\pm$0.44&1.00$\pm$0.089\\
Cui~\etal\cite{cui2019semi}   &0.829(0.847)$\pm$0.045*&11.00&3.97$\pm$0.38&1.03$\pm$0.089\\
Baur~\etal\cite{baur2017semi}  &0.778(0.795)$\pm$0.071*&4.15&4.06$\pm$0.40&1.05$\pm$0.082\\
\hline
\textbf{ROAM ($\kappa=0)\dagger$}   &\textbf{0.852(0.866)$\pm$0.037}&\textbf{14.05}&\textbf{3.91$\pm$0.35}&\textbf{0.99$\pm$0.067}\\
\textbf{ROAM($\kappa=2$)} &\textbf{0.872(0.881)$\pm$0.024}&\textbf{16.73}&\textbf{3.78$\pm$0.28}&\textbf{1.00$\pm$0.077}\\
\textbf{ROAM$\ddagger$}          &\textbf{0.870(0.873)$\pm$0.023}&\textbf{16.50}&\textbf{3.87$\pm$0.31}&\textbf{1.00$\pm$0.061}\\
\hline
Upper Bound             &0.871(0.886)$\pm$0.044*&16.60&3.72$\pm$0.42&0.95$\pm$0.087\\
\textbf{ROAM-UB} &\textbf{0.893(0.902)$\pm$0.024}&\textbf{19.54}&\textbf{3.56$\pm$0.34}&\textbf{0.91$\pm$0.075}\\
\hline 
\end{tabular}}
\end{table}
\subsubsection{Structures Level Results}
\label{SecStrctResults}
\begin{figure}[t]
\centering
\resizebox{0.5\textwidth}{!}{
\includegraphics[width=0.50\textwidth]{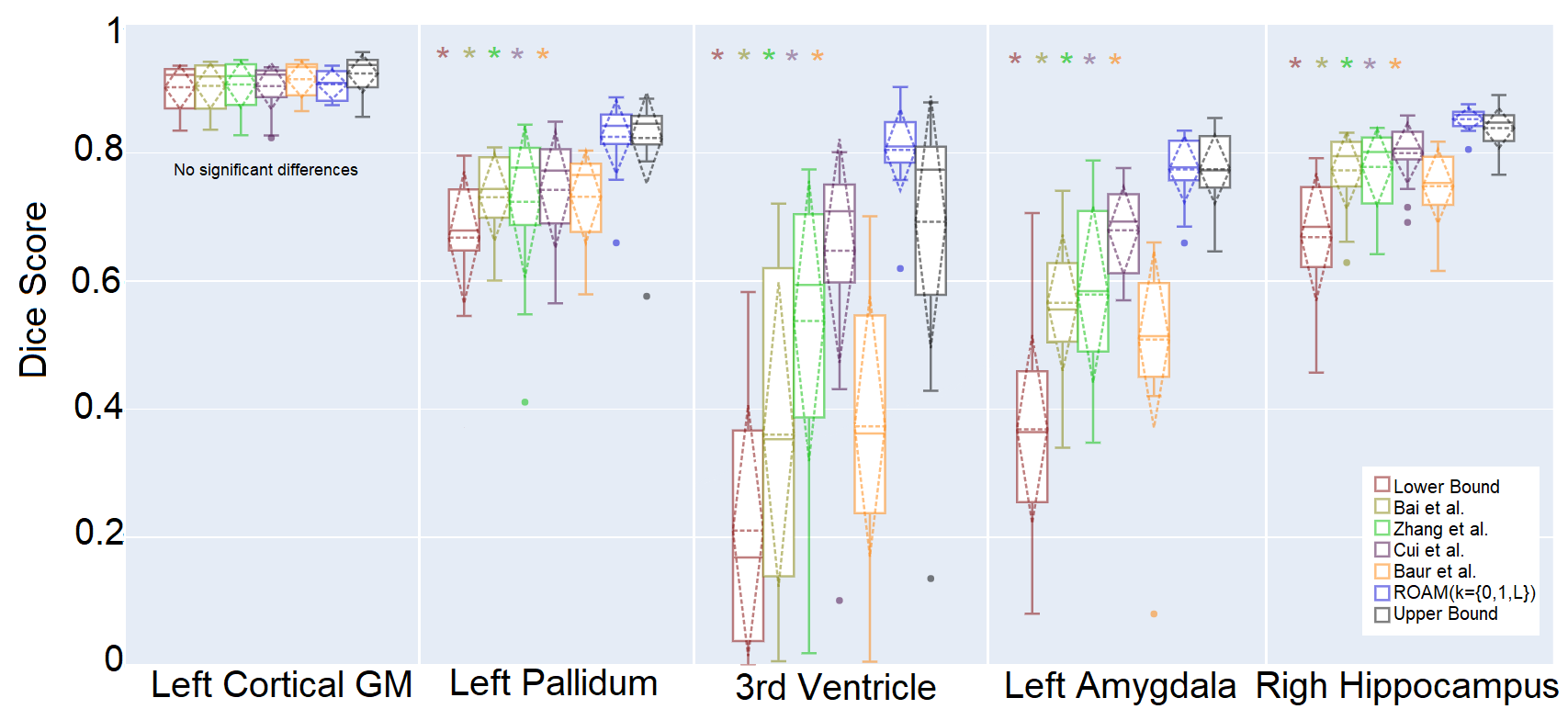}}
\caption{Dice score for selected structures. Our method significantly outperforms all other SSL methods in most structures.} 
\label{fig2}
\end{figure}

In Fig.\ref{fig2} we report the performance at the internal brain structures. It can be seen that our method significantly outperforms all other SSL methods in most structures, and even outperforms the upper bound in the Right Hippocampus and 3rd Ventricle.
Besides, the performance of our method is consistent across different structures, this clearly shown in the Left Pallidum, 3rd Ventricle, Left Amygdala, and Right Hippocampus. Despite the fact our model achieves lower performance in Left Cortical GM, yet the difference is not statistically significant.
\subsubsection{Qualitative Results}
\label{SecQuaResults}
\begin{figure*}[t]
\centering
\includegraphics [width=\textwidth]{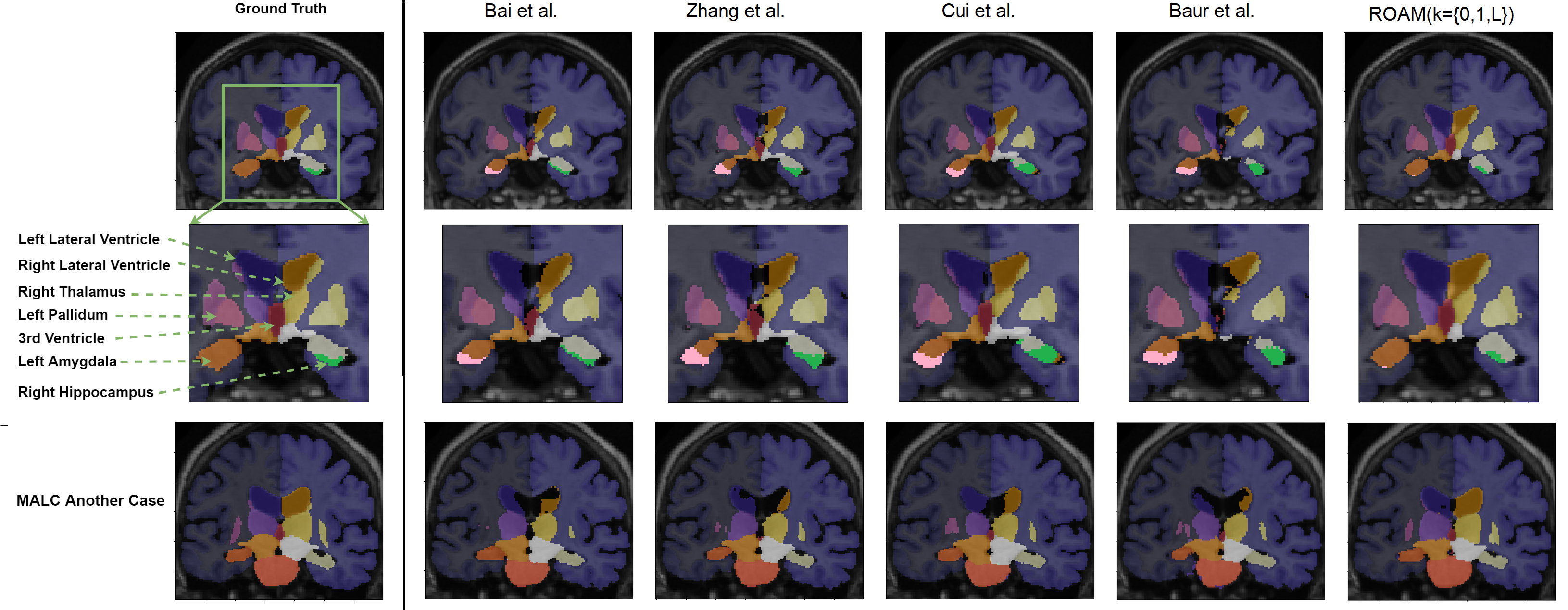}
\caption{Qualitative results of different SSL methods. First row: Coronal view of Whole brain segmentation. Second row: Cropped version highlighting some selected structures. Third Row: Another case from MALC dataset. The results show the superiority of ROAM against all methods.} 
\label{fig4}
\end{figure*}

To provide more insights on the performance, we generate the segmentation predictions for ROAM in addition to all other SSL methods. The results in the previous sections are verified by the segmentation predictions shown in Fig.\ref{fig4}. The first row represents the whole brain segmentation for the SSL methods on the MALC dataset. The second row shows a cropped version highlighting some selected structures. In the cropped version, for example, we highlighted right and left lateral ventricle, right thalamus, right hippocampus, left palladium, left amygdala, and 3rd ventricle. Despite the complexity of these small structures, ROAM performs more reliable than all SSL methods. To support our findings, we also include another case from the MALC dataset in the third row. 
In general, ROAM predicts more accurate results than other SSL methods indicating its generalization ability to other domains. 
\begin{figure}[t]
\centering
\resizebox{0.40\textwidth}{!}{
\includegraphics[width=0.45\textwidth]{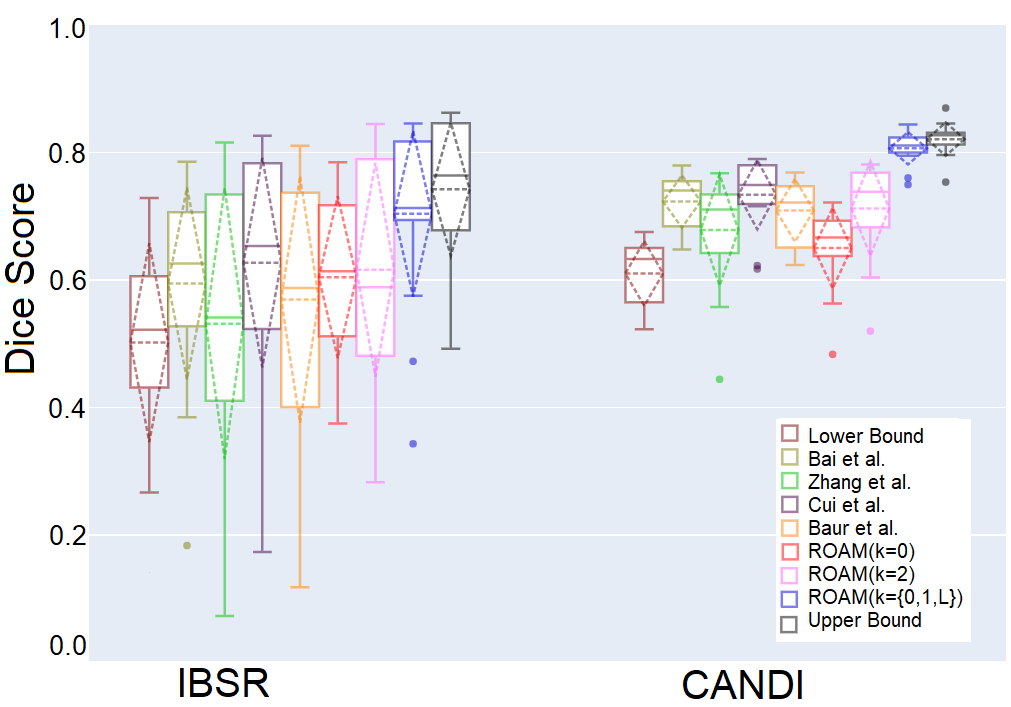}}
\caption{Domain shift results. ROAM performs less sensitive to the domain shift problem compared with other models}
\label{fig3}
\end{figure}
\subsubsection{Comparison with SOTA for Whole Brain Segmentation}
\label{SecFSResults}
To realize the effectiveness of ROAM in a fully-supervised fashion, we run our method on the labeled data alone. Yet, several changes have been done to achieve this experiment. First, the labeled batch is mixed with its permuted version. Second, no sharpening nor pseudo labeling steps are used. Third, we propagate on the supervised loss $\mathcal{L}_{CE}$. In this experiment, we train our model on the whole MACL training set (15 volumes) for 80 epochs and report the results on the testing set (15 volumes) at the last epoch. Finally, we compare our method with U-net, and QuickNAT; \cite{roy2019quicknat}. In contrast to our models and Unet, which are only trained on the MALC training set, QuickNAT is pre-trained using 581 labeled volumes from IXI dataset\footnote{\url{http://brain-development.org/ixi-dataset/}}, and further fine-tuned on the MALC training dataset (15 volumes). Table \ref{table2} shows that all the variations of ROAM significantly outperform Unet and on par and sometimes outperform QuickNAT without a sophisticated pre-training mechanism. Note that ROAM ($\kappa=0$) is a special case of our method where the mixup is performed at the input space \emph{\ie\ MixMatch}. Further, the results show that our models achieve lower standard deviations compared to other methods. In this experiment, we show that our simple but elegant ROAM operation leads to SOTA results without the need for large datasets.
\begin{table}[!t]
\centering
\caption{Dice score for fully supervised models. ROAM significantly outperforms both Unet and QiuckNAT without sophisticated pre-training mechanism}
\resizebox{0.48\textwidth}{!}{
\begin{tabular}{l l r}
\hline
Model Name&Mean(median)$\pm$std& RI(\%) \\
\hline
Unet &0.874(0.888)$\pm$0.039&0\\
QuickNAT &0.895(N/A)$\pm$0.055&2.40\\
\hline
ROAM ($\kappa=0$)                  &0.890(0.898)$\pm$0.025&1.83\\
ROAM ($\kappa=\{0,1,L\}$)                 &\textbf{0.895(0.901)$\pm$0.022}&2.40\\
ROAM ($\kappa=2$)                      &\textbf{0.897(0.906)$\pm$0.025}&2.63\\
\hline
\end{tabular}}
\label{table2}
\end{table}
\subsubsection{Realistic evaluation of ROAM} 
\label{SecRE}
\begin{figure}[h]
\centering
\resizebox{0.5\textwidth}{!}{
\includegraphics[width=0.5\textwidth]{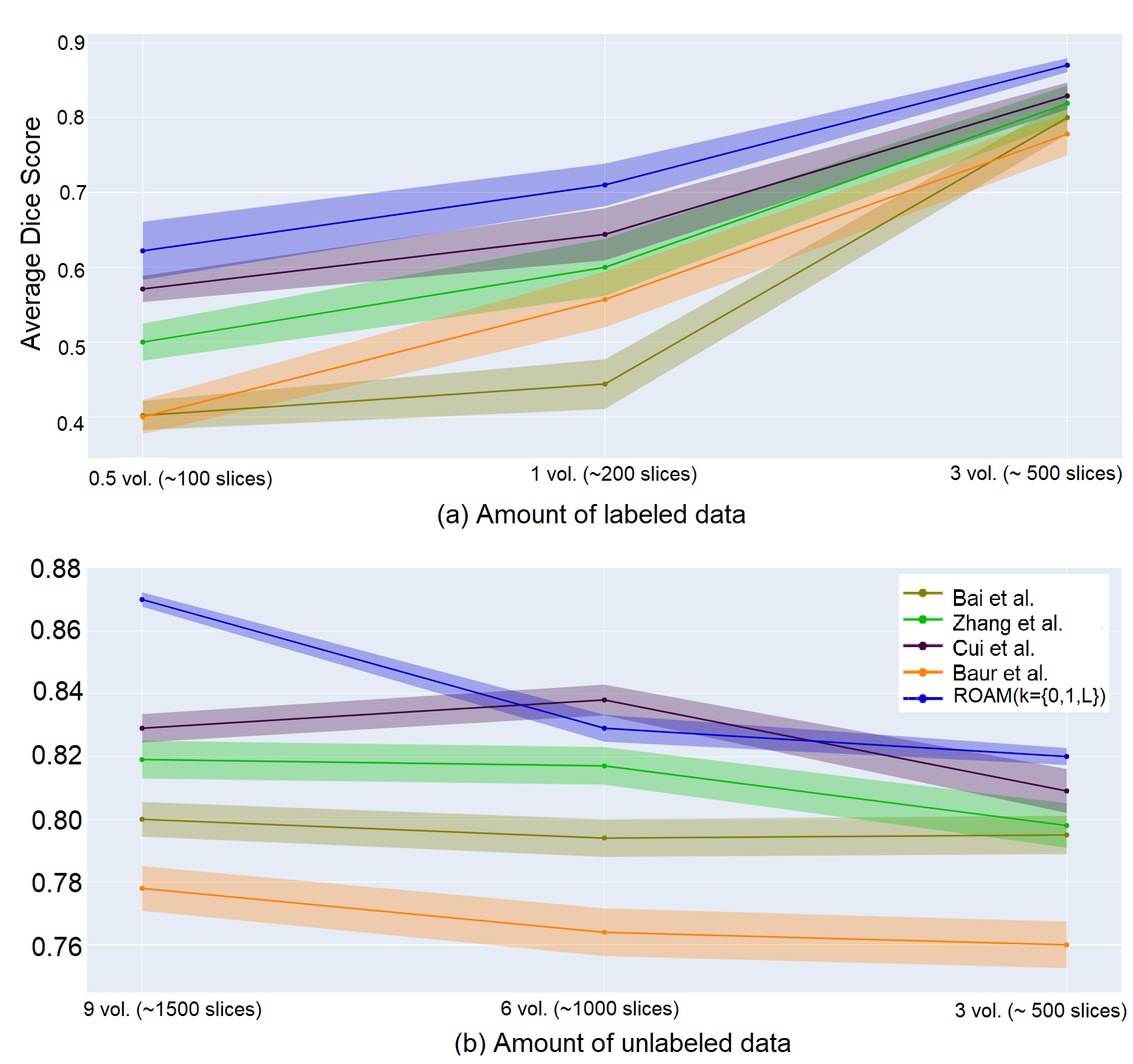}}
\caption{Varying amount of data. The shaded region represent the standard deviation. The  more labeled or unlabeled data being used,  the  higher  performance  and  confidence  of  our model  comparing  to  others.}
\label{fig7}
\end{figure}

\paragraph{Changing amount of labeled data} 
\label{secChangAmountLabeled}
At first, we fix the number of unlabeled data at 1500 slices while changing the amount of labeled data from 100 to 500 slices. With successive increases in the amount of the labeled, the higher performance and confidence of our model compared to others (\cf Fig.~\ref{fig7}.a). Yet, the confidence level is inconsistent in other models. 
\paragraph{Changing amount of unlabeled data.} 
\label{secChangAmountunlabeled}
In this experiment, we fix the number of the labeled data at 500 slices while reducing the amount of the unlabeled from 1500 to 500 slices. The results are shown in Fig.~\ref{fig7}.b. In contrast to other methods, our model shows its superior \emph{w.r.t} variable amount of unlabeled data.
Yet, Cui~\etal\cite{cui2019semi} achieves insignificant higher Dice at 1000 unlabeled slices.
 \subsubsection{Domain shift results} 
 \label{secCrosD} 
 Moreover, we test all models in the presence of the domain shift. The trained models picked and tested on IBSR and CANDI datasets. The results in Fig.\ref{fig3} show a drastic drop in all models, including the baseline ones. This drop is higher on the ISBR dataset. However, ROAM($\kappa=\{0,1, L\}$) performs just well in both cases and less sensitive to the domain shift problem compared with other models, including its variants. Surprisingly, although ROAM($\kappa=2$) achieves one of the best results on MALC dataset, yet it has less generalization ability than ROAM($\kappa=\{0,1, L\}$).

%% file: discussion.tex
\section{Discussion}
\label{Discussion}
In this paper, we propose a semi-supervised medical imaging segmentation framework that utilizes the modern regularization methods to boost the model with newly generated data points. Our method inspired by the work of MixMatch which has been proposed in the computer vision domain for classification tasks. Yet, our method overcomes the limitations of MixMatch by introducing a \textit{random layer mixup} \ie {ROAM} at the input and hidden spaces that suits the complexity of medical images segmentation. 
\subsubsection{ROAM Performance Across Different Datasets} We validate our method using three publicly available datasets for the brain images. The data is heterogeneous however the structures in the brain images almost rigid and geometry constrained. However ROAM performs pretty well across different datasets and outperform all SSL methods. The robustness of ROAM has been shown in the brain segmentation, in which, ROAM always obtains the best results. 
Surprisingly, ROAM-LB consistently achieves better results than many SSL methods, which gives an indication that this simple regularization technique, with just a few labeled data, works better than many of SSL methods that have access to a large amount of unlabeled data. Intuitively, ROAM generates new data points through its linear interpolation. The effectiveness of this operation is essential at a lower data regime where more data needed for the training.
\subsubsection{Generalizability \& Domain Mismatch} One way to alleviate the need for a large amount of annotated data is to utilize datasets coming from different sources. Oftentimes, these datasets come with many challenges, i.e. different cohorts, scanning protocols, and scanners. This leads to a technical challenge, so-called domain shift. We investigate ROAM under these conditions and have noticed that all SSL methods, including ROAM, suffer in the presence of domain shift. Yet, ROAM was less sensitive to this problem, see Fig.\ref{fig3}. Nevertheless, we make no claim here that our approach is domain agnostic, and further research in handing domain shift in SSL is definitely an interesting direction.  
\subsubsection{Convergence} It has been shown that Manifold mixup is guaranteed to be converged when the mixup performed at a sufficiently deep hidden layer as long as the dimensionality of that layer is greater than the number of the classes \cite{verma2019manifold}. This condition is satisfied in this paper where the dimensionality of the hidden layers $> 32$, which is greater than the number of segmentation classes in our datasets \ie 28 for brain images. %
\subsubsection{Handling Skip Connections} One important question is how to handle the skip connections when mixing at a random layer of the U-Net. Do the skip connections get interpolated using the same lambda as the convolution layers or just forwarded without any mixup?. For example, when mixing two samples $x_{1}$ and $x_{2}$ at a random hidden layer \ie $\kappa=2$, the skip connections which go around that layer will still hold the original data from the first hidden layer. Therefore, they will not correspond to the mixed-up labels properly, and hence, that might lead to serious issues. Fortunately, this problem did not happen in our main scenarios, \ie performing random mixup at $\kappa=\{0, 1, L\}$ because the mixed-labels are corresponding to the mixed data as well. However, it is not the case for maniofld mixup, e.g. $\kappa = 2$, which surprisingly shows interesting results. One of the reasons could be attributed to the choice of the beta distribution parameter, \ie $\alpha$. For instance, when $\alpha$ is less than 1, then the mixed data tend to preserve the original data point, and therefore, performing manifold at the bottleneck or other layers might not have such an expected negative impact. One suggestion to handle this issue in the future as follows. When we perform the mixup at a given layer, then we should mix the skip-connections up to that point with the same lambda and the same example-pairing. We investigate this solution on MALC dataset for the SSL and UB models when $\kappa=2$. The results, reported in Table \ref{table6}, show that ROAM performs differently, and no such approach produces consistent results in the given scenarios. For instance, the skip-connections mixup at SSL settings hurts the results while it almost has no effect or a negligible positive effect at the upper bounds. The issue of handling the skip-connections is an intriguing one which could be usefully explored in further research.
\begin{table}[!t]
\centering
\caption{The results for Whole-brain segmentation at $\kappa=2$ with/out skip-connections mixup. ROAM work better without skip-connections mixup at SSL setting, while it perform just lower at the upper bounds. SK: Skip-Connection Miuxp.}
\resizebox{0.47\textwidth}{!}{
\begin{tabular}{p{0.10\linewidth}p{0.23\linewidth}p{0.05\linewidth}p{0.30\linewidth}}
\hline
Dataset&Model&SK&Mean(median)$\pm$std\\
\hline
\hline
MALC &ROAM-SSL&\checkmark&0.834(0.853)$\pm$0.047\\
                     &&\texttimes&0.872(0.881)$\pm$0.024\\
                     & ROAM-UB  &\checkmark&0.892(0.898)$\pm$0.024\\
                      &   &\texttimes&0.890(0.898)$\pm$0.023\\
\hline
\end{tabular}}
\label{table6}
\end{table}
\subsubsection{Validation Datasets} 
\label{secValidationSet}
It has been shown by \cite{oliver2018realistic} that using small validation datasets leads to inconsistency in the results. The smaller the validation set, the larger the variations in the output, which may not reflect the actual performance of the model. Having said that, \cite{oliver2018realistic} argued that a comparison between different SSL models is possible when the validation set is equal to the training one.  In this paper, we consider this and argue that the reported results reflect the actual performance. 
\subsubsection{Hyper-parameters Tuning} 
\label{secHyperparameters}
Besides the standard ones, ROAM involves a set of hyper-parameters and designing choices that should be selected as well. Although fine-tuning such an amount of parameters is a tedious task, our results presented in Table \ref{table0} show that ROAM outperforms all SSL methods in a wide range of hyper-parameters values, thus, with a little effort, you can achieve SOTA performance. To support our argument, we give the following examples. First, ROAM outperforms all SSL models regardless of the selected random layer $\kappa$. Second, the lowest values of ROAM obtained at $\kappa=$\{3 or 4\} with average Dices equal to 82.5\% and 82.8\%, respectively, which are better than all other SSL methods, except \cite{cui2019semi}. Third, all ROAM variations and design choices, which include the sharpening and concatenation steps, outperform all other SSL models. Fourth, it has been shown that the effect of the newly generated data, when $\beta=0$, boosts the performance without the need for the unlabeled loss. Practically, the number of hyper-parameters can be reduced significantly by fixing $\kappa=\{0, 1, L\}$ and just fine-tuning $\alpha$ and $\beta$, which is the standard case in many SSL methods. 

%% file: conclusion.tex
\section{Conclusion}
\label{sec:coclusion}
In this paper, we introduce ROAM, a\textit{ random layer mixup} for semi-supervised learning in medical images segmentation. We show that our method is less prone to over-fitting and has better generalization property. 
Our experiments show a superior and SOTA performance of our method on the whole brain image segmentation. We tested ROAM in both supervised and semi-supervised settings and we have shown its preference against other approaches. 
Our comprehensive experiments show that our method utilizes both labeled and unlabeled data efficiently, proofing its stability, superiority, and consistency. 
So far, the quality of the pseudo labels mainly depends on the initial guess and the mixup coefficient $\lambda$, however, one could think of modeling this coefficient as a function of uncertainty measures. Also, to generate more realistic mixed-up data, one could think of performing the mixup operation on disentangled representations.
Our experiment demonstrates a robust performance of our method under domain shift. Nevertheless, domain invariant SSL methods should be further investigated.
ROAM, as the other SSL methods, can be affected by the imbalance-classes datasets. Instead of naive mixup, one can investigate more intelligent ways of data mixing. 